\documentclass[twocolumn,twoside]{IEEEtran}

\usepackage{amsmath,amstext,amssymb,epsf}
\usepackage{epsfig}
\usepackage{verbatim}

\numberwithin{equation}{section} 

\newcommand{\NID}{ \textsc {NID} }
\newcommand{\NGD}{ \textsc {NGD} }

\newcommand{\NCD}{ \textsc {NCD} }

\newcommand{\SVM}{ \textsc {SVM} }

 \newtheorem{lemma}{Lemma}[section]
 \newtheorem{theorem}[lemma]{Theorem}

 \newtheorem{definition}[lemma]{Definition}
 \newtheorem{rem}[lemma]{Remark}
\newenvironment{remark}{\begin{rem}}{\hspace*{\fill}$\diamondsuit$\end{rem}}
 \newtheorem{ex}[lemma]{Example}

\date{}
\begin{document}
\title{The Google Similarity Distance}
\author{Rudi L. Cilibrasi
and Paul M.B. Vit\'anyi
\thanks{
The material of this paper was presented in part at the
IEEE ITSOC Information Theory Workshop 2005 on Coding and Complexity, 
29th Aug. - 1st Sept., 2005, Rotorua, New Zealand, and the
IEEE Intn'l Symp. Information Theory, Seattle, Wash. USA, August 2006.
Manuscript received April 12, 2005;
final revision June 18, 2006. 
Rudi Cilibrasi was supported in part by the Netherlands
 BSIK/BRICKS project,
and by NWO project 612.55.002.
He is at the Centre for Mathematics and Computer 
Science (Centrum voor Wiskunde en Informatica),
Amsterdam, the Netherlands.
Address:
CWI, Kruislaan 413,
1098 SJ Amsterdam, The Netherlands.
Email: {\tt Rudi.Cilibrasi@cwi.nl}.
Paul Vitanyi's work was done in part while the author was on sabbatical leave
at National ICT of Australia, Sydney Laboratory at UNSW.
He is affiliated with the Centre for Mathematics and Computer 
Science (Centrum voor Wiskunde en Informatica)
and the University of Amsterdam, both in Amsterdam, the Netherlands.
Supported in part
by the EU  EU Project RESQ IST-2001-37559,
the ESF QiT Programmme,
the EU NoE PASCAL, and the Netherlands BSIK/BRICKS project.
Address:
CWI, Kruislaan 413,
1098 SJ Amsterdam, The Netherlands.
Email: {\tt Paul.Vitanyi@cwi.nl}.}
}

\maketitle

\begin{abstract}
Words and phrases acquire meaning from the way they are used
in society, from their relative semantics to other words and
phrases. For computers the equivalent of `society' is `database,' and the
equivalent of `use'
is `way to search the database.'
We present a new theory of similarity between words
and phrases 
based on information distance and Kolmogorov complexity.
To fix thoughts we use the world-wide-web as database, and Google
as search engine. The method is also applicable to other
search engines and databases.
This theory is then applied to construct a method to 
automatically extract similarity, the Google similarity distance,
of words and phrases from the world-wide-web using Google
page counts.
The world-wide-web is the largest database on earth,
and the context information
entered by millions of independent users
averages out to provide automatic semantics of useful quality.
We give applications in hierarchical clustering,
classification, and language translation.
We give examples
to distinguish between colors and numbers,
cluster names of paintings by 17th century Dutch masters and
names of books by English novelists,
the ability to understand
emergencies, and primes,
and we demonstrate the ability to do a simple
automatic English-Spanish translation.
Finally, we use the WordNet database as an objective baseline against which to judge
the performance of our method.  We conduct a massive randomized trial
in binary classification using support vector machines to learn categories based on
our Google distance, resulting in an a mean agreement of 87\% with
the expert crafted 
WordNet categories. 

{\em Index Terms}---

accuracy comparison with WordNet categories,
automatic classification and clustering,
automatic meaning discovery using Google, 
automatic relative semantics,
automatic translation,
dissimilarity semantic distance,
Google search,
Google distribution via page hit counts,
Google code,
Kolmogorov complexity,
normalized compression distance ( \NCD ),
normalized information distance ( \NID ),
normalized Google distance ( \NGD ),
meaning of words and phrases extracted from the web,
parameter-free data-mining,
universal similarity metric

\end{abstract}

\section{Introduction}
\label{sect.intro}

Objects can be given literally, like the literal
four-letter genome of a mouse,
or the literal text of {\em War and Peace} by Tolstoy. For
simplicity we take it that all meaning of the object
is represented by the literal object itself. Objects can also be
given by name, like ``the four-letter genome of a mouse,''
or ``the text of {\em War and Peace} by Tolstoy.'' There are
also objects that cannot be given literally, but only by name,
and that acquire their meaning from their contexts in background common
knowledge in humankind, like ``home'' or ``red.''
To make computers more intelligent one would like
to represent meaning in computer-digestable form.
Long-term and labor-intensive efforts like
the {\em Cyc} project \cite{cyc:intro} and the {\em WordNet}
project \cite{wordnet} try to establish semantic relations
between common objects, or, more precisely, {\em names} for those
objects. The idea is to create
a semantic web of such vast proportions that rudimentary intelligence,
and knowledge about the real world, spontaneously emerge.
This comes at the great cost of designing structures capable
of manipulating knowledge, and entering high
quality contents in these structures
by knowledgeable human experts. While the efforts are long-running
and large scale, the overall information entered is minute compared
to what is available on the world-wide-web.

The rise of the world-wide-web has enticed millions of users
to type in trillions of characters to create billions of web pages of
on average low quality contents. The sheer mass of the information
about almost every conceivable topic makes it likely
that extremes will cancel and the majority or average is meaningful
in a low-quality approximate sense. We devise a general
method to tap the amorphous low-grade knowledge available for free
on the world-wide-web, typed in by local users aiming at personal
gratification of diverse objectives, and yet globally achieving
what is effectively the largest semantic electronic database in the world.
Moreover, this database is available for all by using any search engine
that can return aggregate page-count estimates for a large
range of search-queries, like Google.

Previously, we and others developed a compression-based method to
establish a universal similarity metric among objects
given as finite binary strings 
\cite{BGLVZ,LBCKKZ01,malivitch:simmet,cidervit:mus,civit:cbc,KLR04,CBVA06},
which was widely reported \cite{NS03,TRN03,PlS04}. Such objects
can be genomes, music pieces in MIDI format, computer programs
in Ruby or C, pictures in simple bitmap formats, or time sequences such as
heart rhythm data. This method is feature-free in the sense
that it doesn't analyze the files looking for particular
features; rather it analyzes all features simultaneously
and determines the similarity between every pair of objects
according to the most dominant shared feature. The crucial
point is that the method analyzes the objects themselves.
This precludes comparison of abstract notions or other objects
that don't lend themselves to direct analysis, like
emotions, colors, Socrates, Plato, Mike Bonanno and Albert Einstein.
While the previous method that compares the objects themselves is
particularly suited to obtain knowledge about the similarity of
objects themselves, irrespective of common beliefs about such
similarities, here we develop a method that uses only the name
of an object and obtains knowledge about the similarity of objects,
a quantified relative Google semantics,
by tapping available information generated by multitudes of
web users. 
Here we are reminded of the words of D.H. Rumsfeld \cite{Ru01}
``A trained ape can know an awful lot /
Of what is going on in this world /
Just by punching on his mouse /
For a relatively modest cost!''
In this paper, the Google semantics of a word or phrase
consists of the set of web pages returned by the query concerned.

\subsection{An Example:}
While the theory we propose is rather intricate, the resulting method
is simple enough. We give an example:
At the time of doing the experiment, a Google search 
for ``horse'', returned 
46,700,000 hits. The number of hits for the
search term ``rider'' was 12,200,000. Searching 
for the pages where both ``horse'' and ``rider'' occur gave
 2,630,000 hits, and
Google indexed 8,058,044,651 web pages.
Using these numbers in the main formula  \eqref{eq.NGD} we derive below, with
$N=8,058,044,651$, this yields a Normalized Google Distance
 between the terms ``horse''
and ``rider'' as follows: 
\[
\NGD(horse,rider)
\approx 0.443.
\]
In the sequel of the paper we argue that the \NGD is a normed semantic
distance between the terms in question, 
usually (but not always, see below) 
in between 0 (identical) and 1 (unrelated), in
the cognitive space invoked by the usage of the terms
on the world-wide-web as filtered by Google. Because of the vastness
and diversity of the web this may be taken as related 
to the current use
of the terms in society. 
We did the same calculation when Google indexed only one-half
of the number of pages: 4,285,199,774. It is instructive that the
probabilities of the used search terms didn't change significantly over
this doubling of pages, with number of hits for ``horse''
equal 23,700,000, for ``rider'' equal 6,270,000, and
for ``horse, rider'' equal to 1,180,000.
The $\NGD(horse,rider)$ we computed
in that situation was $\approx 0.460$. This is in line with our contention
that the relative frequencies of web pages containing
search terms gives objective information about the semantic
relations between the search terms. If this is the case, then
the Google probabilities of search terms and the computed \NGD's
should stabilize (become scale invariant) with a growing Google database.

\subsection{Related Work:}
There is a great deal of work in both cognitive psychology \cite{LD97},
linguistics, and computer science, about using word (phrases)
frequencies in text corpora to develop measures for word similarity
or word association, partially surveyed in \cite{TC03,TKS02},
going back to at least
\cite{Le69}. One of the most successful is Latent Semantic Analysis
(LSA) \cite{LD97} that has been applied in various forms in a great
number of applications. We discuss LSA and its relation to the present
approach in Appendix~\ref{app.LSA}. 
As with LSA, many
other previous approaches of extracting corollations from text documents are based
on text corpora that are many order of magnitudes smaller, and that are
in local storage,  and on
assumptions that are more refined, than what we propose. 
In contrast, \cite{CS04,BbA05} and the many references cited there,
use the web and Google counts to identify lexico-syntactic patterns or other data.
Again, the theory, aim, feature analysis,
 and execution are different from ours, and cannot
meaningfully be compared. Essentially, our method below
automatically extracts semantic relations between arbitrary objects
from the web in a manner 
that is feature-free,
up to the search-engine used, and computationally feasible. 
This seems to be a new direction altogether.

\subsection{Outline:}
The main thrust is to develop a new theory of semantic
distance between a pair of objects, based on (and unavoidably biased by) a
background contents
consisting of a database of documents. An example of the latter
is the set of pages constituting the world-wide-web.
Similarity relations between pairs of objects is distilled from the documents
by just using the number of documents in which the objects occur, singly
and jointly (irrespective
of location or multiplicity). 
For us, the Google semantics of a word or phrase
consists of the set of web pages returned by the query concerned.
Note that this can mean that terms with different meaning have the
same semantics, and that opposites like "true" and "false" often have
a similar semantics. Thus, we just discover associations between
terms, suggesting a likely relationship. As the web grows, the Google
semantics may become less primitive. 
The theoretical underpinning 
is based 
on the theory of Kolmogorov complexity
\cite{liminvit:kolmbook}, and is in terms of coding and compression.
This allows to express and prove properties
of absolute relations between objects that cannot even
be expressed by other
approaches. 
The theory, application, and
the particular \NGD formula to express the bilateral semantic relations are
(as far as we know) not equivalent to any earlier
theory, application, and formula in this area.
The current paper is a next step
in a decade of cumulative research in
this area, of which the main thread is \cite{liminvit:kolmbook,
BGLVZ,Li01,malivitch:simmet,cidervit:mus,civit:cbc} 
with \cite{LBCKKZ01,BLM03} using
the related approach of \cite{LiVi96}.
We first start with a technical introduction outlining 
some notions underpinning our
approach: Kolmogorov
complexity, information distance, and compression-based
similarity metric (Section~\ref{sect.tech}).
Then we give a technical description of the Google distribution,
the Normalized Google Distance, and the universality of these
notions (Section~\ref{sect.google}).
While it may be possible in principle that other methods can use
the entire world-wide-web to determine semantic similarity between terms,
we do not know of a method that both uses the entire web, or computationally
can use the entire web, and (or) has
the same aims as our method. To validate our method we therefore cannot
compare its performance to other existing methods. Ours is a new proposal
for a new task. We validate the method
in the following way: by theoretical analysis, by anecdotical evidence in a plethora
of applications, and by systematic and massive comparison of accuracy in
a classification application compared to the
uncontroversial body of knowledge in the WordNet database. 
In Section~\ref{sect.google} we give the theoretic underpinning of the
method and prove its universality. 
In Section~\ref{sect.exp} we present a plethora of clustering and
classification experiments to validate the universality, robustness, and
accuracy of our proposal. In Section~\ref{sect.validation}
we test repetitive automatic performance against uncontroversial semantic knowledge:
We present the results of a massive randomized classification
trial we conducted to gauge the accuracy of our method to the expert knowledge
as implemented over the decades in the WordNet database.
The preliminary publication \cite{CV04} of this work on the web archives
was widely reported and discussed, for example \cite{NS05,Sl05}.
The actual experimental data can be downloaded from
\cite{Ci04}. The method is implemented as an easy-to-use software
tool available on the web \cite{Ci03}, available to all.

\subsection{Materials and Methods:}
\label{sect.materials}
The application of the theory we develop is a method that is justified
by the vastness of the world-wide-web, the assumption that the mass of information
is so diverse that the frequencies of pages returned by Google queries  averages
the semantic information in such a way that one can distill a valid
semantic distance between the query subjects. 
It appears to be the only method that starts from scratch, 
is feature-free in that it uses
just the web and a search engine to supply contents,
and automatically generates relative semantics between words and phrases. 
A possible drawback of our method is that it relies
on the accuracy of the returned counts. As noted in \cite{BbA05},
the returned google counts are inaccurate, 
and especially if one uses the boolean
OR operator between search terms, at the time of writing.
The AND operator we use is less problematic, and we do not use the OR
operator. Furthermore, 
Google apparently estimates the number of hits based on samples, and the number
of indexed pages changes rapidly. To compensate for the latter effect, we have 
inserted a normalizing mechanism in the CompLearn software. Generally though,
if search engines have peculiar ways of counting number of hits,
in large part this should not matter, as long as some
reasonable conditions hold on how counts are reported. Linguists judge
the accuracy of Google counts trustworthy enough:
In \cite{KL05} (see also the many references to related research)
 it is shown that web searches for rare two-word phrases 
correlated well with the frequency found in traditional corpora, 
as well as with human judgments of whether those phrases were natural.
Thus, Google is the simplest means to get the most information.
Note, however, that a single Google query takes a fraction of a second,
and that Google restricts every IP address to a maximum 
of (currently) 500 queries
per day---although they are cooperative enough 
to extend this quotum for noncommercial
purposes.
The experimental evidence provided here shows that the combination of Google
and our method yields
reasonable results,
 gauged against common sense (`colors' are different from `numbers')
and against the expert knowledge in the WordNet data base.
A reviewer suggested  downscaling
our method by testing it on smaller text corpora. This does not seem useful.
Clearly perfomance will deteriorate with decreasing data base size.
 A thought experiment using
the extreme case of a single web page consisting of a single term
suffices. Practically addressing this issue is begging the question.
 Instead, 
in Section~\ref{sect.google} we theoretically analyze the 
relative semantics of search terms established 
using all of the web, and its universality with
respect to the relative semantics of search terms
using subsets of web pages.

\section{Technical Preliminaries}
\label{sect.tech}
The basis of much of the theory explored in this paper is
Kolmogorov complexity. For an introduction and details
see the textbook \cite{liminvit:kolmbook}. Here we give some intuition and
notation. We assume a fixed reference universal programming system.
Such a system may be a general computer language like LISP
or Ruby, and it may also be a fixed reference universal Turing
machine in a given standard enumeration of Turing machines.
The latter choice has the advantage of being formally simple and hence
easy to theoretically manipulate. But the choice makes no difference in
principle, and the theory is invariant under changes among the universal
programming systems, provided we stick to a particular choice.
We only consider universal programming systems such that the
associated set of programs is a prefix code---as is the case in
all standard computer languages.
The {\em Kolmogorov complexity} of a string $x$ is the length, in bits, of the
shortest computer program of the fixed reference
computing system that produces $x$ as output.  The choice of computing system
changes the value of $K(x)$ by at most an additive fixed constant.
Since $K(x)$ goes to infinity with $x$, this additive fixed constant
is an ignorable quantity if we consider large $x$.
One way to think about the Kolmogorov complexity $K(x)$ is to view it
as the length, in bits, of the ultimate compressed version  from
which $x$ can be recovered by a general decompression program.
Compressing $x$ using the compressor {\em gzip} results
in a file $x_g$ with (for files that contain redundancies) the length $|x_g| < |x|$.
Using a better compressor {\em bzip2} results in a file
$x_b$ with (for redundant files) usually $|x_b| < |x_g|$;
using a still better compressor like {\em PPMZ} results in
a file $x_p$ with (for again appropriately redundant files) $|x_p| < |x_b|$.
The Kolmogorov complexity $K(x)$ gives a lower bound on the ultimate value:
for every existing compressor, or compressors that are possible but not known,
we have that $K(x)$ is less or equal to the length of the compressed version of $x$.
That is, $K(x)$ gives us the ultimate value of the length of a compressed
version of $x$ (more precisely, from which version $x$ can be reconstructed
by a general purpose decompresser), and our task in designing better and better
compressors is to approach this lower bound as closely as possible.

\subsection{Normalized Information Distance:}
In \cite{BGLVZ} we considered the following 
notion: given two strings $x$ and $y$,
 what is the length of the shortest
binary program  in the reference universal computing system such that
the program computes output $y$ from input $x$,
and also output $x$ from input $y$.
This is called the {\em information distance}
and denoted as $E(x,y)$.
It turns out that, up to a negligible logarithmic additive term,
\[
E(x,y)= K(x,y) - \min \{K(x),K(y) \},
\]
where $K(x,y)$ is the binary length
of the shortest program that produces the pair $x,y$ and
a way to tell them apart.
This distance $E(x,y)$ is actually a metric: up to close precision
we have $E(x,x)=0$, $E(x,y)>0$ for $x \neq y$,
 $E(x,y)=E(y,x)$ and $E(x,y) \leq E(x,z)+E(z,y)$,
for all $x,y,z$. We now consider a large class of {\em admissible
distances}: all distances (not necessarily metric) that are nonnegative,
symmetric, and {\em computable} in the sense that for every such distance $D$
there is a prefix program that,
given two strings $x$ and $y$, has binary length equal to
the distance $D(x,y)$ between $x$ and $y$.
Then,
\begin{equation}\label{eq.minor}
E(x,y) \leq D(x,y) + c_D,
\end{equation}
where $c_D$ is a constant that depends only on $D$ but not on $x,y$, and
we say that $E(x,y)$ {\em minorizes} $D(x,y)$ up to an additive constant.
We call the information distance $E$ {\em universal} for the family
of computable distances, since the former minorizes every member
of the latter family up to an additive constant.
If two strings $x$ and $y$ are close
according to {\em some} computable distance $D$, then they are at
least as close according to distance $E$. Since every feature in which
we can compare two strings can be quantified in terms of a distance,
and every distance can be viewed as expressing a quantification of
how much of a particular  feature the strings do not have in common
 (the feature being quantified by that distance),
the information distance determines the distance between two
strings minorizing the {\em dominant} feature in which they are similar.
This means that, if we consider more than two strings, 
the information distance between every pair may be based
on minorizing a different dominating feature.
If small strings differ by an information distance which is
large compared to their sizes, then the strings are very different.
However, if two very large strings differ by the same (now relatively small)
information distance, then they are very similar.
Therefore, the information distance itself is not suitable
to express true similarity. For that we must define
a relative information distance: we need to normalize the information distance.
Such an approach was first proposed in \cite{LBCKKZ01} in the context
of genomics-based phylogeny, and improved
in \cite{malivitch:simmet} to the one we use here.
The {\em normalized information distance (\NID)}
has  values between 0 and 1,
and it inherits the universality of the information distance in the sense
that it minorizes, up to a vanishing additive term, every
other possible normalized computable distance (suitably defined).
In the same way as before we can identify the computable normalized
distances with computable similarities according to some features, and
the \NID discovers for every pair of strings the feature in which they
are most similar, and expresses that similarity on a scale from 0 to 1
(0 being the same and 1 being completely different in the sense
of sharing no features). Considering a set of strings, the 
feature in which two strings are most similar may be a different
one for different pairs of strings. The \NID is defined by

\begin{equation}\label{eq.nid}
 \NID(x,y) = \frac{K(x,y) - \min(K(x),K(y))}{\max(K(x),K(y))}.
\end{equation}

It has several wonderful properties that justify its description as the
most informative metric \cite{malivitch:simmet}.
\subsection{Normalized Compression Distance:}
The \NID is
uncomputable since the Kolmogorov complexity is uncomputable.
But we can use real data compression programs to approximate the Kolmogorov
complexities $K(x),K(y),K(x,y)$.
A compression algorithm defines a computable function from strings to
the lengths of the compressed versions of those strings. Therefore,
the number of bits of the compressed version of a
string is an upper bound on Kolmogorov complexity of that string,
up to an additive constant depending on the compressor but not
on the string in question. Thus, if $C$ is a compressor and we use $C(x)$
to denote the length of the compressed version of a string $x$,
then we arrive at the {\em Normalized Compression Distance}:
\begin{equation}\label{eq.ncd}
 \NCD(x,y) = \frac{C(xy) - \min(C(x),C(y))}{\max(C(x),C(y))},
\end{equation}
where for convenience we have replaced the pair $(x,y)$ in the formula
by the concatenation $xy$. This transition raises several tricky
problems, for example how the \NCD approximates the \NID
if $C$ approximates $K$, see \cite{civit:cbc},
which do not need to concern us here. Thus, the
\NCD is actually a family of compression functions parameterized 
by the given data
compressor $C$.  The \NID is the limiting case, where $K(x)$
denotes the number of bits in the shortest code for $x$ from
which $x$ can be decompressed by a general purpose computable
decompressor.

\section{Theory of Googling for Similarity}
\label{sect.google}
Every text corpus or particular user combined
with a frequency extractor
defines its own relative frequencies of
words and phrases usage. In the world-wide-web and 
Google setting there are
millions of users and text corpora, 
each with its own distribution. 
In the sequel, we show (and prove) that the Google distribution
is universal for all the individual web users distributions.
The number of web pages currently indexed by Google is approaching
$10^{10}$. Every common search term occurs in millions of web pages.
This number is so vast, and the number of web authors generating
web pages is so enormous (and can be assumed to be a
truly representative very large sample from humankind),
that the probabilities of Google search terms, conceived as
the frequencies of page counts returned by Google divided by
the number of pages indexed by Google,
approximate the actual relative frequencies of those search terms
as actually used in society. Based on this premise,
the theory we develop in this paper states 
that the relations represented by the 
Normalized Google Distance \eqref{eq.NGD}
approximately capture
the assumed true semantic
relations governing the search terms.
The \NGD formula \eqref{eq.NGD}
 only uses the probabilities of search terms
extracted from the text corpus in question. We use the world-wide-web
and Google, but the same method may be used with other text corpora
like the King James version of the 
Bible or the Oxford English Dictionary and frequency count
extractors, or the world-wide-web again and Yahoo as frequency
count extractor. In these cases one obtains a text corpus and
frequency extractor biased semantics of the search terms.
To obtain the true relative frequencies of words and phrases
in society is a major problem in applied linguistic research.
This requires analyzing representative random samples of sufficient
sizes. The question of how to sample randomly and
representatively is a continuous source of debate.
Our contention that the web is such a large and diverse text corpus,
and Google such an able extractor, that the relative 
page counts approximate the true societal word- and phrases
usage, starts to be supported by current real linguistics research
\cite{Ec04,KL05}.

\subsection{The Google Distribution:}
Let the set of singleton {\em Google search terms}
be denoted by ${\cal S}$. 
In the sequel we use both singleton
search terms and doubleton search terms $\{\{x,y\}: x,y \in {\cal S} \}$.
Let the set of web pages indexed (possible of being returned)
by Google be $\Omega$. The cardinality of $\Omega$ is denoted
by $M=|\Omega|$, and at the time of this writing 
$8\cdot 10^9 \leq M \leq 9 \cdot 10^9$
(and presumably greater by the time of reading this).
Assume that a priori all web pages are equi-probable, with the probability
of being returned by Google being $1/M$.  A subset of $\Omega$
is called an {\em event}. Every {\em  search term} $x$ usable by Google
defines a {\em singleton Google event} ${\bf x} \subseteq \Omega$ of web pages
that contain an occurrence of $x$ and are returned by Google
if we do a search for $x$.
Let $L: \Omega \rightarrow [0,1]$ be the uniform mass probability
function.
The probability of
an event ${\bf x}$ is $L({\bf x})=|{\bf x}|/M$.
 Similarly, the {\em doubleton Google event} ${\bf x} \bigcap {\bf y}
\subseteq \Omega$ is the set of web pages returned by Google
if we do a search for pages containing both search term $x$ and
search term $y$.
The probability of this event is $L({\bf x} \bigcap {\bf y})
= |{\bf x} \bigcap {\bf y}|/M$.
We can also define the other Boolean combinations: $\neg {\bf x}=
\Omega \backslash {\bf x}$ and ${\bf x} \bigcup {\bf y} =
\neg ( \neg {\bf x} \bigcap \neg {\bf y})$, each such event
having a probability equal to its cardinality divided by $M$.
If ${\bf e}$ is an event obtained from the basic events ${\bf x}, {\bf y},
\ldots$, corresponding to basic search terms $x,y, \ldots$,
by finitely many applications of the Boolean operations,
then the probability $L({\bf e}) = |{\bf e}|/M$.

\subsection{Google Semantics:}
Google events capture in a particular sense
all background knowledge about the search terms concerned available
(to Google) on the web. 
\begin{quote}
The Google event ${\bf x}$, consisting of the set of 
all web pages containing one or more occurrences of the search term $x$,
thus embodies, in every possible sense, all direct context
in which $x$ occurs on the web. This constitutes the Google semantics
of the term. 
\end{quote}
\begin{remark}
\rm
It is of course possible that parts of
this direct contextual material link to other web pages in which $x$ does not
occur and thereby supply additional context. In our approach this indirect
context is ignored. Nonetheless, indirect context may be important and
future refinements of the method may take it into account. 
\end{remark}
\subsection{The Google Code:}
The event ${\bf x}$ consists of all
possible direct knowledge on the web regarding $x$.
Therefore, it is natural
to consider code words for those events
as coding this background knowledge. However,
we cannot use the probability of the events directly to determine
a prefix code, or, rather the underlying information content implied
by the probability.
The reason is that
the events overlap and hence the summed probability exceeds 1.
By the Kraft inequality \cite{CT91} this prevents a
corresponding set of code-word lengths.
The solution is to normalize:
We use the probability of the Google events to define a probability
mass function over the set $\{\{x,y\}: x,y \in {\cal S}\}$
of  Google search terms, both singleton and doubleton terms. There are
$|{\cal S}|$ singleton terms, and 
${ |{\cal S}|  \choose 2}$  doubletons consisting of a pair of non-identical
terms. 
Define
\[
 N= \sum_{\{x,y\} \subseteq {\cal S}} |{\bf x} \bigcap
{\bf y}|,
\]
counting each singleton set and each doubleton set (by definition
unordered) once in the summation. Note that this means that
for every pair $\{x,y\} \subseteq {\cal S}$, with $x \neq y$,
the web pages $z \in {\bf x} \bigcap
{\bf y}$ 
 are counted three times: once in ${\bf x} =  {\bf x} \bigcap
{\bf x}$, once in ${\bf y} =  {\bf y} \bigcap
{\bf y}$, and
once in  ${\bf x} \bigcap
{\bf y}$.
Since every web page that is indexed by Google contains at least
one occurrence of a search term, we have $N \geq M$. On the other hand,
web pages contain on average not more than a certain constant $\alpha$
search terms. Therefore, $N \leq \alpha M$.
Define
\begin{align}\label{eq.gpmf}
g(x) = g(x,x), \; \;
g(x,y) =  L({\bf x} \bigcap {\bf y}) M/N =|{\bf x} \bigcap {\bf y}|/N.
\end{align}
Then, $\sum_{\{x,y\} \subseteq {\cal S}} g(x,y) = 1$.
This $g$-distribution changes over time,
and between different samplings
from the distribution. But let us imagine that $g$ holds
in the sense of an instantaneous snapshot. The real situation
will be an approximation of this.
Given the Google machinery, these are absolute probabilities
which allow us to define the associated prefix code-word lengths (information contents)
 for
both the singletons and the doubletons.
The {\em Google code} $G$
is defined by
\begin{align}\label{eq.gcc}
G(x)= G(x,x), \; \;
G(x,y)= \log 1/g(x,y) .
\end{align}
\subsection{The Google Similarity Distance:}
In contrast to strings $x$ where the complexity $C(x)$ represents
the length of the compressed version of $x$ using compressor $C$, for a search
term $x$ (just the name for an object rather than the object itself),
the Google code of length $G(x)$ represents the shortest expected
prefix-code word length of the associated Google event ${\bf x}$.
The expectation
is taken over the Google distribution $g$.
In this sense we can use the Google distribution as a compressor
for the Google semantics associated with the search terms.
The associated \NCD, now called the
{\em normalized Google distance (\NGD)} is then defined
by \eqref{eq.NGD}, and can be rewritten as the right-hand expression:
\begin{eqnarray}\label{eq.NGD}
 \NGD(x,y)& =&\frac{G(x,y) - \min(G(x),G(y))}{\max(G(x),G(y))}
\\&=&  \frac{  \max \{\log f(x), \log f(y)\}  - \log f(x,y) }{
\log N - \min\{\log f(x), \log f(y) \}},
\nonumber
\end{eqnarray}
where $f(x)$ denotes the number of pages containing $x$, and $f(x,y)$
denotes the number of pages containing both $x$ and $y$, as reported by Google.
This $\NGD$ is an approximation to the $\NID$ of \eqref{eq.nid}
using the prefix code-word lengths (Google code)
generated by the Google distribution as defining a compressor
approximating the length of the Kolmogorov code, using
the background knowledge on the web as viewed by Google
as conditional information. In practice, use the page counts
returned by Google for the frequencies, and we have to
choose $N$.  From the right-hand side term in \eqref{eq.NGD}
it is apparent that by increasing $N$ we decrease the \NGD , everything gets 
closer together, and
by decreasing $N$ we increase the \NGD , everything gets further apart.
Our experiments suggest that every reasonable
($M$ or a value greater than any $f(x)$) value can be used as
normalizing factor  $N$,
and our
results seem  in general insensitive to this choice.  In our software, this
parameter $N$ can be adjusted as appropriate, and we often use $M$ for $N$.
The following are the main properties
of the \NGD (as long as we choose parameter $N \geq M$):
\begin{enumerate}
\item
The {\em range} of the \NGD is in between 0 and $\infty$ (sometimes
slightly negative if the Google counts are untrustworthy and state
$f(x,y) > \max \{f(x),f(y)\}$, See Section~\ref{sect.materials});
\begin{enumerate}
\item
If  $x=y$ or if $x \neq y$ but frequency $ f(x)=f(y)= f(x,y)>0$,
then $\NGD(x,y)=0$. That is, the semantics of $x$ and $y$ in
the Google sense is the same.
\item
If frequency $f(x)=0$,
then for every search term $y$ we have $f(x,y)=0$,
and the $\NGD(x,y)= \infty / \infty$,
which we take to be 1 by definition.
\end{enumerate}
\item
The $\NGD$ is always nonnegative and $\NGD(x,x)=0$ for every $x$.
For every pair $x,y$ we have $\NGD(x,y)=\NGD(y,x)$: it is symmetric.
However, the \NGD is {\em not a metric}: it does not satisfy
$\NGD(x,y) > 0$ for every $x \neq y$. As before, let ${\bf x}$ denote 
the set of web pages
containing one or more occurrences of $x$. For example, choose $x \neq y$
with ${\bf x}={\bf y}$. Then,
$f(x)=f(y)=f(x,y)$ and $\NGD(x,y)=0$.
Nor does the \NGD satisfy the triangle inequality
$\NGD(x,y) \leq \NGD(x,z)+\NGD(z,y)$ for all $x,y,z$. For example,
choose ${\bf z}= {\bf x} \bigcup {\bf y}$,
${\bf x} \bigcap {\bf y}= \emptyset$,
${\bf x}={\bf x} \bigcap {\bf z}$,
${\bf y}={\bf y} \bigcap {\bf z}$,
and $|{\bf x}|=|{\bf y}|= \sqrt{N}$.
Then, $f(x)=f(y)=f(x,z)=f(y,z)=\sqrt{N}$, $f(z)=2\sqrt{N}$, and $f(x,y)=0$.
This yields $\NGD(x,y)=\infty$ and
$\NGD(x,z)=\NGD(z,y)=2/ \log N$, which violates the triangle
inequality for all $N$.
\item
The \NGD is {\em scale-invariant} in the following sense: Assume that when
the number $N$ of pages indexed by Google (accounting
for the multiplicity of different search terms per page)  
grows,
the number of pages containing a given search term
goes to a fixed fraction of $N$, and so does the number of pages
containing a given conjunction of search terms.
This means that if $N$ doubles,
then so do the $f$-frequencies. For the \NGD to give us an objective
semantic relation between search terms,
it needs to become stable when the number $N$ grows unboundedly.
\end{enumerate}


\subsection{Universality of Google Distribution:}
A central notion in the application of compression to learning
is the notion of ``universal distribution,'' see \cite{liminvit:kolmbook}.
Consider an effective enumeration ${\cal P} = p_1,p_2, \ldots$ 
of probability mass functions with
domain ${\cal S}$. The list ${\cal P}$ can be finite or countably
infinite.
\begin{definition}\label{def.unimult}
\rm
A probability mass function $p_u$ occurring in ${\cal P}$
is {\em universal} for ${\cal P}$, if for every $p_i$ in  ${\cal P}$
there is a constant $c_i >0$ and $\sum_{i \neq u} c_i \geq 1$,
 such that for every $x \in {\cal S}$ we have
$ p_u(x) \geq  c_i \cdot p_i (x)$. 
Here $c_i$ may depend on the indexes $u,i$,
but not on the functional mappings of the elements
of list ${\cal P}$ nor  on $x$.
\end{definition}

If $p_u$ is universal for ${\cal P}$, then it
immediately follows that for every $p_i$ in ${\cal P}$,
the prefix code-word length for source word $x$, see
\cite{CT91}, 
associated with $p_u$, minorizes the prefix code-word
length associated with $p_i$,
by satisfying  $\log 1/p_u(x) \leq \log 1/p_i(x) + \log 1/c_i$, 
for every $x \in {\cal S}$.

In the following we consider partitions of the set of web pages, 
each subset in the partition together with a
probability mass function of search terms.
For example, we may consider
the list ${\cal A} =  1,2, \ldots ,a$ of
{\em web authors producing pages} on the web, and consider
the set of web pages produced by each web author, or some other partition.
``Web author'' is just a metaphor we use for convenience.
Let web author $i$ of the list  ${\cal A}$ produce  the set of
web pages $\Omega_i$  and denote $M_i=|\Omega_i|$.
We identify a web author $i$ with the set of web
pages $\Omega_i$ he produces.
Since we have no knowledge of the set of web authors, we consider
every possible partion of $\Omega$ into one
of more equivalence classes,
 $\Omega = \Omega_1 \bigcup \cdots \bigcup \Omega_a$,
$\Omega_i \bigcap \Omega_j = \emptyset$ 
($1 \leq i \neq j \leq a \leq |\Omega|$),
as defining a  realizable set of web authors ${\cal A}= 1, \ldots, a$.

Consider a partition of $\Omega$ into 
$\Omega_1, \ldots , \Omega_a$.
A search term $x$ usable by Google defines an event
${\bf x}_i \subseteq \Omega_i$ of web pages produced by web author $i$
that contain search term $x$. Similarly, ${\bf x}_i \bigcap {\bf y}_i$
is the set of web pages produced by $i$ that is returned by Google
searching for pages containing both search term $x$ and
search term $y$.
Let
\[
 N_i = \sum_{\{x,y\} \subseteq {\cal S}} |{\bf x}_i \bigcap
{\bf y}_i|.
\]
Note that there is an $\alpha_i \geq 1$ such that $M_i \leq N_i \leq \alpha_i M_i$.
For every search term $x \in {\cal S}$ define
a probability mass function $g_i$,
the {\em individual web author's Google distribution},
 on the sample space $\{\{x,y\}: x,y \in S\}$ by
\begin{align}\label{eq.gpmfi}
g_i(x) = g_i(x,x), \; \;
g_i(x,y) =  |{\bf x}_i \bigcap {\bf y}_i|/N_i .
\end{align}
Then, $ \sum_{\{x,y\} \subseteq {\cal S}}
g_i(x,y) = 1$.
\begin{theorem}
Let $\Omega_1, \ldots , \Omega_a$ be any partition of $\Omega$
into subsets (web authors), and let $g_1, \ldots , g_a$
be the corresponding individual Google distributions.
Then the Google distribution $g$ is universal for the enumeration 
$g,g_1, \ldots , g_a$.
\end{theorem}
\begin{proof}
We can express the overall Google distribution in terms
of the individual web author's distributions:
\[
g(x,y) = \sum_{i \in {\cal A}} \frac{N_i}{N} g_i (x,y).
\]
Consequently, $g(x,y) \geq (N_i/N) g_i(x,y)$. Since also $g(x,y) \geq g(x,y)$,
we have shown that $g(x,y)$ is universal for the family $g,g_1, \ldots , g_a$
of individual web author's google distributions, according to
Definition~\ref{def.unimult}.
\end{proof}

\begin{remark}
\rm
Let us show that, for example,
the uniform distribution $L(x)=1/s$ ($s=|{\cal S}|$) 
over the search terms $x \in {\cal S}$ is
not universal, for $s > 2$.
By the requirement $\sum c_i \geq 1$, the sum taken over 
the number $a$ of web authors in the list ${\cal A}$, there is
an $i$ such that $c_i \geq 1/a$. Taking the uniform distribution on 
say $s$ search terms assigns probability $1/s$ to each of them.
By the definition of
universality of a probability mass function for the list of individual
Google probability mass functions $g_i$, we can choose the function
$g_i$ freely (as long as $a \geq 2$, and there is another function $g_j$
to exchange probabilities of search terms with). So choose some 
search term $x$ and set $g_i(x)=1$, and $g_i (y)=0$ for all search
terms $y \neq x$. Then, we obtain 
$g(x)=1/s \geq c_i g_i (x) \geq 1/a$. This yields the required
contradiction for $s > a \geq 2$.
\end{remark}

\subsection{Universality of Normalized Google Distance: }
Every individual web author produces both an individual
Google distribution $g_i$, and an {\em individual prefix code-word length} $G_i$
associated with $g_i$ (see \cite{CT91} for this code)
for the search terms. 
\begin{definition}
\rm
The associated {\em individual normalized Google 
distance} $\NGD_i$ of web author $i$ is defined
according to \eqref{eq.NGD}, with $G_i$ substituted for $G$.
\end{definition}
These Google distances $\NGD_i$ can be viewed as 
the individual semantic distances according
to the bias of web author $i$. These individual semantics are
subsumed in the general Google semantics in the following sense:
The normalized Google distance is {\em universal} for the
family of individual normalized Google distances,
in the sense that it is as about as small as the least
individual normalized Google distance, with high probability.
Hence the Google semantics as evoked by all of the web society
in a certain sense captures the biases or knowledge of
the individual web authors.
In Theorem~\ref{theo.ngd} we show that, for every $k \geq 1$,
the inequality
\begin{equation}\label{eq.uni}
\NGD(x,y) < \beta \;  \NGD_i(x,y) +  \gamma,
\end{equation}
with 
\begin{align*}
\beta &= \frac{\max\{G_i(x),G_i(y)\}}{ \max\{G(x),G(y)\}}
\leq 1+ \frac{\log (2k)}{\max\{G(x),G(y)\}}
\\ \gamma &= \frac{\min\{G_i(x),G_i(y)\}-
\min\{G(x),G(y)\}+\log N/N_i}{\max\{G(x),G(y)\}} 
\\& \leq \frac{\log (2kN/N_i)}{\max\{G(x),G(y) \}},
\end{align*}
is satisfied with  $g_i$-probability going to 1 with growing $k$.

\begin{remark}
\rm
To interpret \eqref{eq.uni}, we observe that in case $G(x)$ and $G(y)$
are large with respect to $\log k$, then $\beta \approx 1$.
If moreover $\log N/N_i$ is large with respect to $\log k$, then
approximately $\gamma \leq (\log N/N_i)/ \max \{G(x),G(y)\}$.
Let us estimate $\gamma$ for this case under reasonable assumptions.
Without loss of generality assume $G(x) \geq G(y)$.
If $f(x)= |{\bf x}|$, the number of pages returned on query $x$,
then $G(x)= \log (N/f(x))$. Thus, approximately 
$\gamma \leq (\log N/N_i)/
(\log N/f(x))$. The uniform expectation of
$N_i$ is $N/|{\cal A}|$, and $N$ divided by that
expectation of $N_i$ equals $|{\cal A}|$,
the number of web authors producing web pages.
The uniform expectation of $f(x)$ is $N/|{\cal S}|$, and $N$ divided
by that expectation of $f(x)$ equals $|{\cal S}|$,
the number of Google search terms we use.
Thus, approximately, $\gamma \leq (\log |{\cal A}|)/(\log |{\cal S}|)$,
and the more the number of search terms exceeds the number of web authors,
the more $\gamma$ goes to 0 in expectation.
\end{remark}
\begin{remark}
\rm
To understand \eqref{eq.uni}, we may consider the codelengths involved as
the Google database changes over time.  It is reasonable to expect that both
the total number of pages as well as the total number of search terms in
the Google database will continue to grow for some time.  In this period,
the sum total probability mass will be carved up into increasingly smaller
pieces for more and more search terms.  The maximum singleton and doubleton
codelengths within the Google database will grow.
But the universality property of the Google distribution implies
that the Google distribution's code length for almost all
 particular search terms
will only exceed the best codelength among any of
the individual web authors as in \eqref{eq.uni}.  
The size of this gap will grow more slowly than
the codelength for any particular search term over time.  Thus, the coding
space that is suboptimal in the Google distribution's code is an ever-smaller
piece (in terms of proportion) of the total coding space.
\end{remark}

\begin{theorem}\label{theo.ngd}
For every web author $i \in {\cal A}$,
the $g_i$-probability concentrated on the pairs of search
terms for which \eqref{eq.uni} holds is at least $(1-1/k)^2$.
\end{theorem}
\begin{proof}
The prefix code-word lengths $G_i$ associated with $g_i$ satisfy
$G(x) \leq G_i (x) + \log N/N_i$ and $G(x,y) \leq G_i (x,y) + \log N/N_i$.
Substituting $G(x,y)$ by $G_i (x,y) + \log N/N_i$ in 
the middle term of \eqref{eq.NGD},
we obtain
\begin{equation}\label{eq.uni1}
\NGD(x,y) \leq \frac{ G_i(x,y) - \min \{G(x),G(y)\}+ \log N/N_i}
{\max \{G(x),G(y)\}}.
\end{equation}
{\em Markov's Inequality}
says the following:
Let $p$ be any probability mass function; let $f$
be any nonnegative function with $p$-expected value
${\bf E} = \sum_i p(i) f(i) < \infty$.
For ${\bf E}   >  0$ we have
$\sum_i  \{ p(i): f(i)/{\bf E} > k \}  < 1/k$.

Fix web author $i \in {\cal A}$.
We consider the conditional probability mass functions
$g'(x)=g(x | x \in {\cal S})$ and $g'_i(x)=g_i(x | x \in {\cal S})$
over singleton search terms in ${\cal S}$ (no doubletons):
The $g'_i$-expected
value of $g' (x)/g'_i(x)$ is
$$
\sum_x  g'_i(x) \frac{g' (x)}{g'_i (x)} \leq 1,
$$
since $g'$ is a probability mass function summing to $\leq 1$.
Then, by Markov's Inequality
\begin{equation}
\sum_x \{ g'_i(x): g' (x)/g'_i(x)  >  k \}   <  \frac{1}{k}
\label{(3.3)}
\end{equation}
Since the probability of an event of a doubleton set of search
terms is not greater than that of an event based on
either of the constituent search terms, and the 
probability of a singleton event conditioned on it being a singleton event
is at least as large as the unconditional probability of that event,  
$2g(x) \geq g'(x) \geq g(x)$ and
$2g_i(x) \geq g'_i (x) \geq g_i(x)$. 
If $g(x) > 2k g_i(x)$, then $g'(x)/g'_i(x) >k$ and
the search terms $x$ satisfy the condition of \eqref{(3.3)}.
Moreover, the probabilities satisfy $g_i(x) \leq g'_i(x)$.
Together, it follows from \eqref{(3.3)} that
$\sum_x \{ g_i(x): g (x)/(2g_i(x))  >  k \}   <  \frac{1}{k}$
and therefore
\[
\sum_x \{ g_i(x): g (x)  \leq  2k g_i (x) \}   >  1  - \frac{1}{k}.
\]
For the $x$'s with $g(x) \leq 2k g_i(x)$ we have
$G_i(x) \leq G (x) + \log (2k)$.
Substitute $G_i(x) - \log (2k)$ for $G (x)$ (there is
$g_i$-probability $\geq 1-1/k$ that $G_i(x)-\log (2k) \leq G(x)$)
and  $G_i(y)-\log (2k) \leq G(y)$ 
in \eqref{eq.uni1}, both in the $\min$-term in the numerator,
and in the $\max$-term in the denominator.
Noting that the two $g_i$-probabilities  $(1-1/k)$
are independent, the total $g_i$-probability that both substitutions
are justified is at least $(1-1/k)^2$. 
\end{proof}

Therefore, the Google normalized distance  minorizes
every normalized compression distance based on a particular
user's generated probabilities of search terms, with high probability
up to an error term that in typical cases is ignorable.

\section{Applications and Experiments}
\label{sect.exp}

\begin{figure*}
\centering
\includegraphics[height=4in,width=7in]{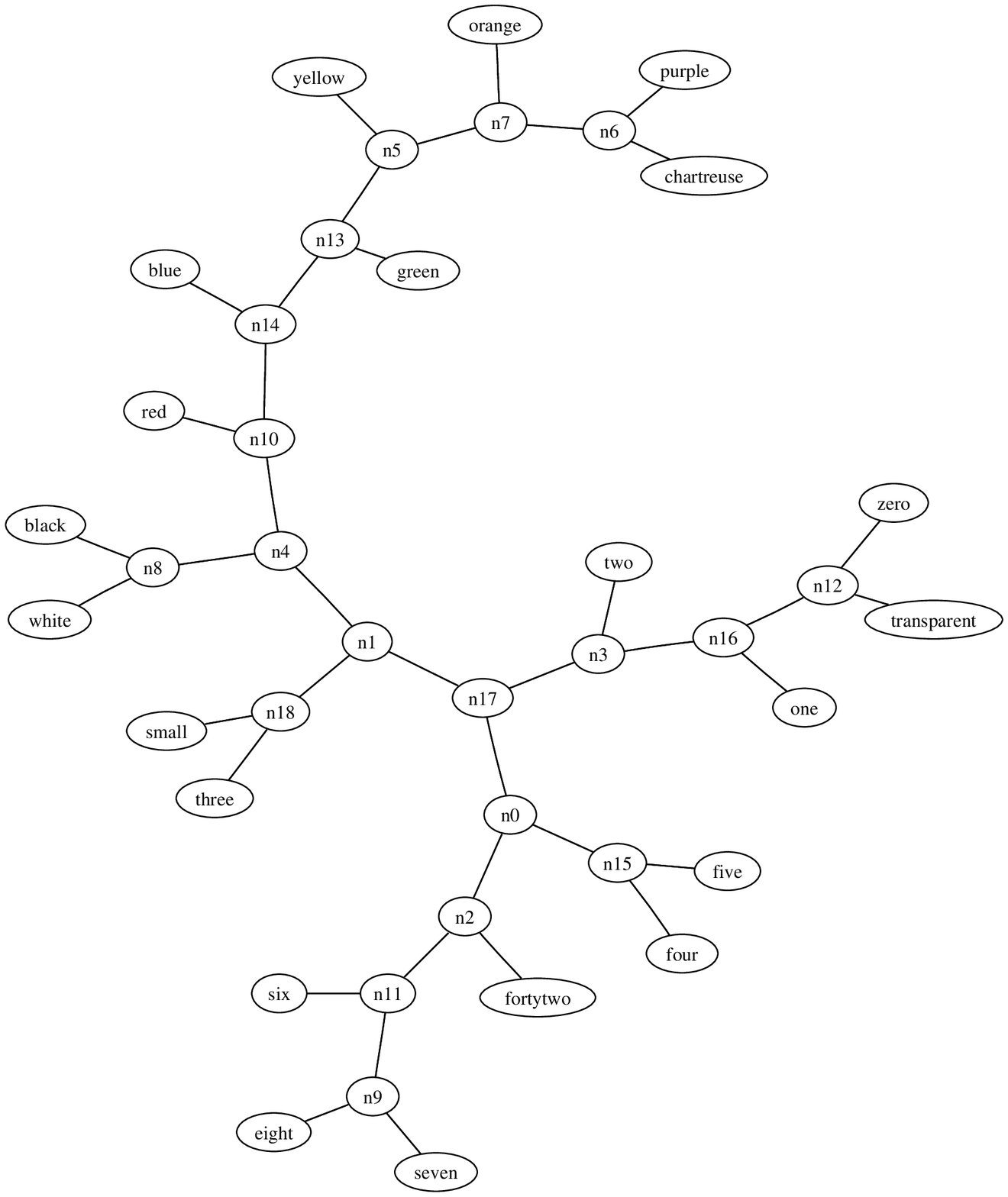}
\caption{Colors and numbers arranged into a tree using \NGD .}
\label{fig.colors}
\end{figure*}

\subsection{Hierarchical Clustering: }
We used our software tool
available from http://www.complearn.org, the same tool that has been
used in our earlier papers \cite{civit:cbc,cidervit:mus}
to construct trees representing hierarchical clusters of objects
in an unsupervised way.  However,
now we use the normalized Google distance (\NGD) instead of the
normalized compression distance (\NCD).  The method
works by first calculating a
distance matrix whose entries are the pairswise \NGD's of the
 terms in the input list.
Then calculate a best-matching unrooted ternary tree using a
novel quartet-method style heuristic based on randomized hill-climbing
using a new fitness objective function for
the candidate trees.
Let us briefly explain what the method does; for more explanation see 
\cite{CV05,civit:cbc}.
Given a set of objects as points in a space provided with
a (not necessarily metric) distance measure,
the associated {\em distance matrix}
has as entries the pairwise distances
between the objects. Regardless of the original space and distance measure,
it is always possible to configure $n$ objects is $n$-dimensional
Euclidean space in such a way that the associated distances are
identical to the original ones, resulting in an identical distance
matrix.
This distance matrix contains the pairwise distance relations
according to the chosen measure in raw form. But in this format
that information is not easily usable, since for $n > 3$ our
cognitive capabilities rapidly fail.
Just as the distance matrix is a reduced form of information
representing the original data set, we now need to reduce the
information even further in order to achieve a cognitively acceptable
format like data clusters.
To extract a hierarchy of clusters
from the distance matrix,
we determine a dendrogram (ternary tree) that agrees
with the distance matrix according to a fidelity measure.
This allows us to extract more information from the data
than just flat clustering (determining disjoint
clusters in dimensional representation).
This method does not just take the strongest link in each case as the ``true'' one,
and ignore all others; instead the tree represents all the relations in the
distance matrix with as little distortion as is possible.
In the particular examples we give below, as in all clustering examples
we did but not depicted, the fidelity was close to 1, meaning that the
relations in the distance matrix are faithfully represented in the tree.
The objects to be clustered are search terms
consisting of the names of colors,
numbers, and some tricky words.  The program automatically organized the colors
towards one side of the tree and the numbers towards the other,
Figure~\ref{fig.colors}.
It arranges the terms which have as only meaning a color or a number, and
\begin{figure*}
\centering
\includegraphics[height=4in,width=7in]{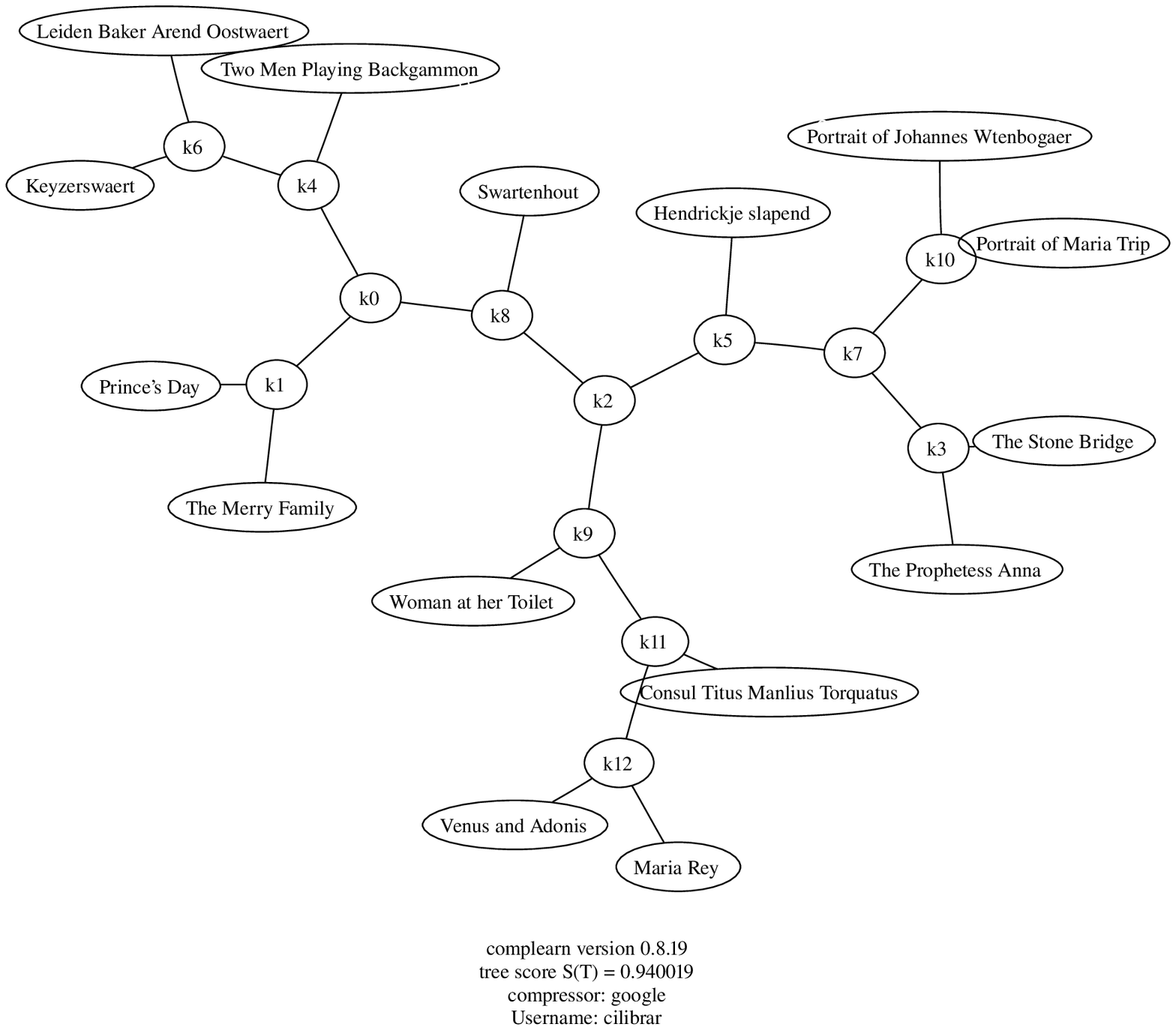}
\caption{Hierarchical clustering of pictures}
\label{fig.painters}
\end{figure*}
nothing else, on the farthest reach of the color side
and the number side, respectively. It puts the
more general terms black and white, and zero, one, and two,
towards the center, thus indicating their
more ambiguous interpretation.  Also, things which were not exactly colors
or numbers are also put towards the center, like the word ``small''.
As far as the authors know there do not exist other experiments that
create this type of semantic distance automatically
from the web using Google or similar search engines. 
Thus, there is no baseline to compare against;
rather the current experiment can be a baseline to evaluate the behavior
of future systems.
\subsection{Dutch 17th Century Painters: }
In the example of  Figure~\ref{fig.painters},
the names of fifteen paintings by Steen, Rembrandt, and Bol
were entered.  
We use the full name as a single Google search term (also in
the next experiment with book titles). 
In the experiment, only painting title names were used; the associated painters
are given below.
We do not know of comparable experiments to use
as baseline to judge the performance; this is a new type of contents 
clustering made possible by the existence of the web and search engines.
 The painters and paintings used are as follows:  

\textbf{ Rembrandt van Rijn:}  
{\em Hendrickje slapend; Portrait of Maria Trip; Portrait of Johannes Wtenbogaert 
; The Stone Bridge ; The Prophetess Anna };

\textbf{ Jan Steen:}  {\em Leiden Baker Arend Oostwaert ; Keyzerswaert ; 
Two Men Playing Backgammon ; Woman at her Toilet ; Prince's Day ; The Merry Family };

\textbf{ Ferdinand Bol:} {\em Maria Rey ; Consul Titus Manlius Torquatus ; 
Swartenhout ; Venus and Adonis }. 


\subsection{English Novelists: }
Another example is English novelists. The authors and texts used are:
\begin{figure*}
\centering
\includegraphics[width=12cm]{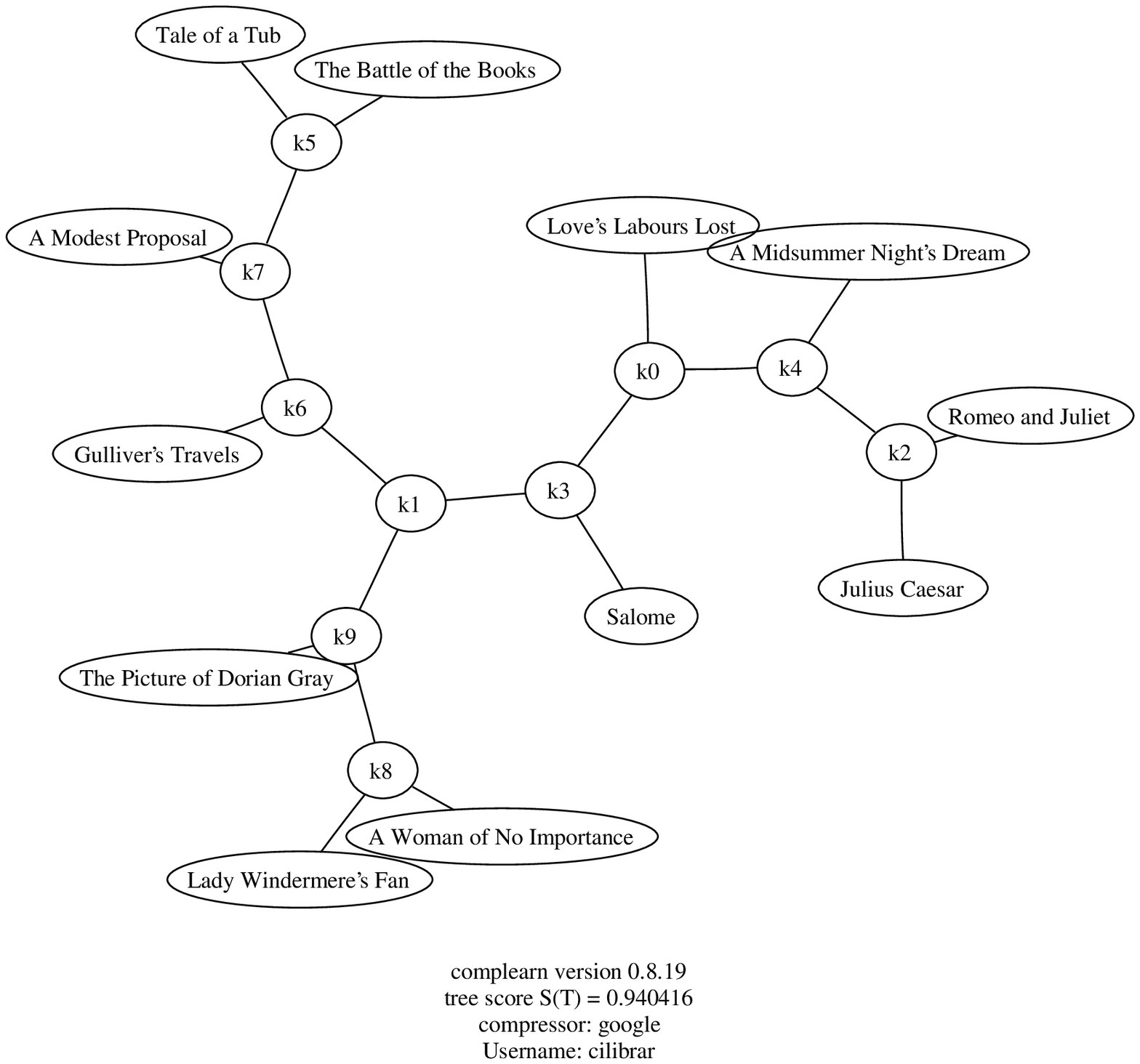}
\caption{Hierarchical clustering of authors}
\label{fig.englishnov}
\end{figure*}

\textbf{William Shakespeare:} {\em A Midsummer Night's Dream; Julius Caesar;
Love's Labours Lost; Romeo and Juliet
}.

\textbf{Jonathan Swift:} {\em The Battle of the Books; Gulliver's Travels;
Tale of a Tub; A Modest Proposal};

\textbf{Oscar Wilde:} {\em Lady Windermere's Fan; A Woman of No Importance;
Salome; The Picture of Dorian Gray}.

The clustering is given in Figure~\ref{fig.englishnov}, and 
to provide a feeling for the figures involved we give the associated
\NGD matrix in Figure~\ref{fig.distmatr}. The $S(T)$ value
in Figure~\ref{fig.englishnov}
gives the fidelity of the tree as a representation of the pairwise
distances in the 
\NGD matrix ($S(T)=1$ is perfect and $S(T)=0$ is as bad as possible. For details
see \cite{Ci03,civit:cbc}). 
\begin{figure*}
\centering
{\footnotesize
\begin{tabular}{ l r r r r r r r r r r r r }
A Woman of No Importance & 0.000 & 0.458 & 0.479 & 0.444 & 0.494 & 0.149 & 0.362 & 0.471 & 0.371 & 0.300 & 0.278 & 0.261 \\
A Midsummer Night's Dream & 0.458 & -0.011 & 0.563 & 0.382 & 0.301 & 0.506 & 0.340 & 0.244 & 0.499 & 0.537 & 0.535 & 0.425 \\
A Modest Proposal & 0.479 & 0.573 & 0.002 & 0.323 & 0.506 & 0.575 & 0.607 & 0.502 & 0.605 & 0.335 & 0.360 & 0.463 \\
Gulliver's Travels & 0.445 & 0.392 & 0.323 & 0.000 & 0.368 & 0.509 & 0.485 & 0.339 & 0.535 & 0.285 & 0.330 & 0.228 \\
Julius Caesar & 0.494 & 0.299 & 0.507 & 0.368 & 0.000 & 0.611 & 0.313 & 0.211 & 0.373 & 0.491 & 0.535 & 0.447 \\
Lady Windermere's Fan & 0.149 & 0.506 & 0.575 & 0.565 & 0.612 & 0.000 & 0.524 & 0.604 & 0.571 & 0.347 & 0.347 & 0.461 \\
Love's Labours Lost & 0.363 & 0.332 & 0.607 & 0.486 & 0.313 & 0.525 & 0.000 & 0.351 & 0.549 & 0.514 & 0.462 & 0.513 \\
Romeo and Juliet & 0.471 & 0.248 & 0.502 & 0.339 & 0.210 & 0.604 & 0.351 & 0.000 & 0.389 & 0.527 & 0.544 & 0.380 \\
Salome & 0.371 & 0.499 & 0.605 & 0.540 & 0.373 & 0.568 & 0.553 & 0.389 & 0.000 & 0.520 & 0.538 & 0.407 \\
Tale of a Tub & 0.300 & 0.537 & 0.335 & 0.284 & 0.492 & 0.347 & 0.514 & 0.527 & 0.524 & 0.000 & 0.160 & 0.421 \\
The Battle of the Books & 0.278 & 0.535 & 0.359 & 0.330 & 0.533 & 0.347 & 0.462 & 0.544 & 0.541 & 0.160 & 0.000 & 0.373 \\
The Picture of Dorian Gray & 0.261 & 0.415 & 0.463 & 0.229 & 0.447 & 0.324 & 0.513 & 0.380 & 0.402 & 0.420 & 0.373 & 0.000 \\
\end{tabular}
}
\caption{Distance matrix of pairwise \NGD's}
\label{fig.distmatr}
\end{figure*}
The question arises why we should expect this. Are names of artistic objects
so distinct? (Yes. The point also being that the distances from every single
object to all other objects are involved. The tree takes this global
aspect into account and therefore disambiguates other meanings of the 
objects to retain the meaning that is relevant for this collection.)
 Is the distinguishing feature subject matter or title style?
In these experiments with objects belonging to the cultural
heritage it is clearly a subject matter. To stress the point we
used ``Julius Caesar'' of Shakespeare. This term occurs on the web
\begin{figure*}
\centering
{ \small
{\large \textbf{Training Data} } \\
\begin{tabular}{l l l l l}
\hline
\\
{\em Positive Training } & (22 cases) \\
avalanche&bomb threat&broken leg&burglary&car collision\\
death threat&fire&flood&gas leak&heart attack\\
hurricane&landslide&murder&overdose&pneumonia\\
rape&roof collapse&sinking ship&stroke&tornado\\
train wreck&trapped miners\\ \\
{\em Negative Training } & (25 cases) \\
arthritis&broken dishwasher&broken toe&cat in tree&contempt of court\\
dandruff&delayed train&dizziness&drunkenness&enumeration\\
flat tire&frog&headache&leaky faucet&littering\\
missing dog&paper cut&practical joke&rain&roof leak\\
sore throat&sunset&truancy&vagrancy&vulgarity\\ \\
{\em Anchors } & (6 dimensions) \\
crime&happy&help&safe&urgent\\
wash\\ \\
\end{tabular}
\begin{tabular}{l l l}\
{\large \textbf{Testing Results} } \\
& Positive tests & Negative tests \\
\hline
Positive&assault, coma,&menopause, prank call, \\
Predictions&electrocution, heat stroke,&pregnancy, traffic jam \\
&homicide, looting,& \\
&meningitis, robbery,& \\
&suicide& \\
\hline
Negative&sprained ankle&acne, annoying sister, \\
Predictions&&campfire, desk, \\
&&mayday, meal \\
{\large \textbf{ Accuracy } } & 15/20 = 75.00\% & \\
\end{tabular}
}
\caption{Google-\SVM learning of ``emergencies.''}
\label{fig.emergencies}
\end{figure*}
\begin{figure*}
\centering
{ \small
{\large \textbf{Training Data} } \\
\begin{tabular}{l l l l l}
\hline
\\
{\em Positive Training } & (21 cases) \\
11&13&17&19&2\\
23&29&3&31&37\\
41&43&47&5&53\\
59&61&67&7&71\\
73\\ \\
{\em Negative Training } & (22 cases) \\
10&12&14&15&16\\
18&20&21&22&24\\
25&26&27&28&30\\
32&33&34&4&6\\
8&9\\ \\
{\em Anchors } & (5 dimensions) \\
composite&number&orange&prime&record\\ \\
\end{tabular}
\begin{tabular}{l l l}\
{\large \textbf{Testing Results} } \\
& Positive tests & Negative tests \\
\hline
Positive&101, 103,&110 \\
Predictions&107, 109,& \\
&79, 83,& \\
&89, 91,& \\
&97& \\
\hline
Negative&&36, 38, \\
Predictions&&40, 42, \\
&&44, 45, \\
&&46, 48, \\
&&49 \\
{\large \textbf{ Accuracy } } & 18/19 = 94.74\% & \\
\end{tabular}
}
\caption{Google-\SVM learning of primes.}
\label{fig.primes}
\end{figure*}
overwhelmingly in other contexts and styles. Yet the collection of
the other objects used, and the semantic distance towards those objects,
given by the \NGD formula,
singled out the semantics of ``Julius Caesar'' relevant to this experiment. 
Term co-occurrence in this specific context of author discussion
is not swamped by other uses of this common English term because
of the particular form of the \NGD and the distances being pairwise.
Using book titles which are common words, like "Horse" and "Rider" by author X,
supposing they exist, this swamping effect will presumably arise.
Does the system gets confused if we add more artists? (Representing
the \NGD matrix in bifurcating trees without distortion
 becomes more difficult for, say, more than 25 objects. See \cite{civit:cbc}.)
 What about other
subjects, like music, sculpture? (Presumably,
the system will be more trustworthy if the subjects are more common
on the web.) These experiments are representative  for 
those we have performed with the current software. 
We did not cherry-pick the best outcomes. For example,
all experiments with these three English writers, with different selections
of four works of each, always yielded a tree so that we could draw a
convex hull around the works of each author, without overlap. 
Interestingly, a similar experiment with Russian authors
gave worse results. 
The readers can do their own experiments to satisfy their 
curiosity using our publicly available
software tool at http://clo.complearn.org/,
 also used in the depicted experiments.
Each experiment can take a long time, hours, because
of the Googling, network traffic, and tree reconstruction and layout.
Don't wait,
just check for the result later.
On the web page
http://clo.complearn.org/clo/listmonths/t.html
the onging cumulated results of all 
(in December 2005 some 160) experiments by the public, 
including the ones depicted here,
are recorded.
\subsection{\SVM -- \NGD Learning: }
We augment the Google method by adding a
trainable component of the learning system. Here we use the Support
Vector Machine (\SVM) as a trainable component.
For the \SVM method
used in this paper, we refer to the exposition 
\cite{burges:svmtut}.
We use LIBSVM software for all of our \SVM experiments.

The setting is a binary classification problem on examples represented
by search terms.  We require a human expert to provide a list of
at least 40 {\em training words},
consisting of at least 20 positive examples and 20 negative examples,
to illustrate the
contemplated concept class.  The expert also provides, say,
six {\em anchor words} $a_1, \ldots , a_6$,
of which half are in some way related to the concept
under consideration.  Then, we use the anchor words to convert each
of the 40 training words $w_1 , \dots , w_{40}$
to 6-dimensional {\em training vectors} $\bar{v}_1 , \ldots , \bar{v}_{40}$.
The entry $v_{j,i}$ of $\bar{v}_j = (v_{j,1}, \ldots , v_{j,6})$
is defined as $v_{j,i} = \NGD (w_j, a_i)$ ($1 \leq j \leq 40$,
$1 \leq i \leq 6$).
The training vectors are
then used to train an \SVM to learn the concept, and then test words
may be classified using the same anchors and trained \SVM model.

In  Figure~\ref{fig.emergencies},
we trained using a list of ``emergencies'' as positive examples,
 and a list of
``almost emergencies'' as negative examples.
The figure is self-explanatory. The accuracy on the test set is 75\%.
In Figure~\ref{fig.primes} the method learns to distinguish prime numbers from
non-prime numbers by example. The accuracy on the test set is about 95\%.
This example illustrates several common features of our
method that distinguish it from the strictly deductive techniques.

\subsection{\NGD Translation:}
Yet another potential application of the \NGD method is in natural
language translation. (In the experiment below we don't use \SVM's
to obtain our result, but determine correlations instead.)
Suppose we are given a system that tries
to infer
a translation-vocabulary among English and Spanish.  Assume that the system
has already determined that there are five words that appear in two different
matched sentences, but the permutation associating the English and Spanish
words is, as yet, undetermined. This setting can arise in real
situations, because English and Spanish have different
rules for word-ordering.  At the outset we
assume a pre-existing vocabulary of eight English words
with their matched Spanish translation.  Can we infer the correct permutation
mapping the unknown words using the pre-existing vocabulary as a basis?
We start by forming an \NGD matrix using the additional English
words of which the translation is known, Figure~\ref{fig.estrans1}.
 We label the columns
 by the translation-known English words, the rows by the
translation-unknown English words. The entries of the matrix are the
\NGD's between the English words labeling the columns and rows.
This constitutes the English basis matrix.
Next, consider the known Spanish
words corresponding to the known English words.
Form a new matrix with the known Spanish words labeling the columns
in the same order as the known English words.
Label the rows of the new matrix by choosing one of the many possible
permutations of the unknown Spanish words.  For each permutation, form the \NGD
matrix for the Spanish words, and compute the pairwise correlation of this
sequence of values to each of the values in the given English word basis
matrix.  Choose the permutation with the highest positive correlation.  If
there is no positive correlation report a failure to extend the vocabulary.
In this example, the computer inferred the correct permutation for the
testing words, see Figure~\ref{fig.estrans2}.
\begin{figure}
\centering
\begin{tabular}{r l}
\textbf{English}&\textbf{Spanish}\\
\hline
tooth & diente \\
joy & alegria \\
tree & arbol \\
electricity & electricidad \\
table & tabla \\
money & dinero \\
sound & sonido \\
music & musica \\
\end{tabular}
\label{fig.estrans1}
\caption{Given starting vocabulary}
\end{figure}
\begin{figure}
\centering
\begin{tabular}{r l}
\textbf{English}&\textbf{Spanish}\\
\hline
plant & bailar \\
car & hablar \\
dance & amigo \\
speak & coche \\
friend & planta \\
\end{tabular}
\caption{Unknown-permutation vocabulary}
\end{figure}
\begin{figure}
\centering
\begin{tabular}{r l}
\textbf{English}&\textbf{Spanish}\\
\hline
plant & planta \\
car & coche \\
dance & bailar \\
speak & hablar \\
friend & amigo \\
\end{tabular}
\caption{Predicted (optimal) permutation}
\label{fig.estrans2}
\end{figure}
\section{Systematic Comparison with WordNet Semantics}
\label{sect.validation}
WordNet \cite{wordnet} is a semantic concordance
of English.  It focusses on the meaning of words by dividing them
into categories.  We use this as follows. A category we want to learn, the concept,
is termed, say,  ``electrical'', and represents
anything that may pertain to electronics.
The negative examples
are constituted by simply everything else.
This category represents a typical expansion of a node in the
WordNet hierarchy. In an experiment we ran,
the accuracy on the test set is 100\%: It turns
out that ``electrical terms'' are unambiguous and easy to learn
and classify by our method. The information in the WordNet database
is entered over the decades by human experts and is precise. The database
is an academic venture and is publicly accessible. Hence it is
a good baseline against which to judge the accuracy of our method in an indirect
manner. While we cannot directly compare the semantic distance, the \NGD, between
objects, we can indirectly judge how accurate it is by using it as basis
for a learning algorithm. In particular, we investigated how
well semantic categories as learned using the \NGD--\SVM approach 
agree with the corresponding WordNet categories.
For details about the structure of WordNet we refer to
the official WordNet documentation available online. We considered 100 
randomly selected semantic categories from the WordNet database.  
For  each
category we  executed the following sequence.
First, the \SVM is trained on 50 labeled training samples. The positive examples
are randomly drawn from the WordNet database in the category in question.
The negative examples are randomly drawn from a dictionary. While the latter examples
may be false negatives, we consider the probability negligible.
Per experiment we used a total of six anchors,
three of which are randomly drawn from the WordNet database category in question,
and three of which are drawn from the dictionary.
Subsequently, every example is converted to
6-dimensional vectors using \NGD. The $i$th entry of the vector is the
\NGD between the $i$th anchor and the example concerned ($1 \leq i \leq 6$).
 The \SVM is trained on the resulting labeled vectors.
The kernel-width and error-cost parameters are automatically
determined using five-fold cross validation.  Finally, testing of how well
the \SVM has learned the classifier is performed
using 20 new examples in a balanced ensemble of positive and negative
examples obtained in the same way, and converted to 6-dimensional vectors
in the same manner, as the training
examples.  This results in an accuracy score of correctly classified 
test examples.
We ran 100 experiments. The actual data are available at
\cite{Ci04}.  
\begin{figure}
\centering
{\tiny
\includegraphics[angle=-90,width=8cm]{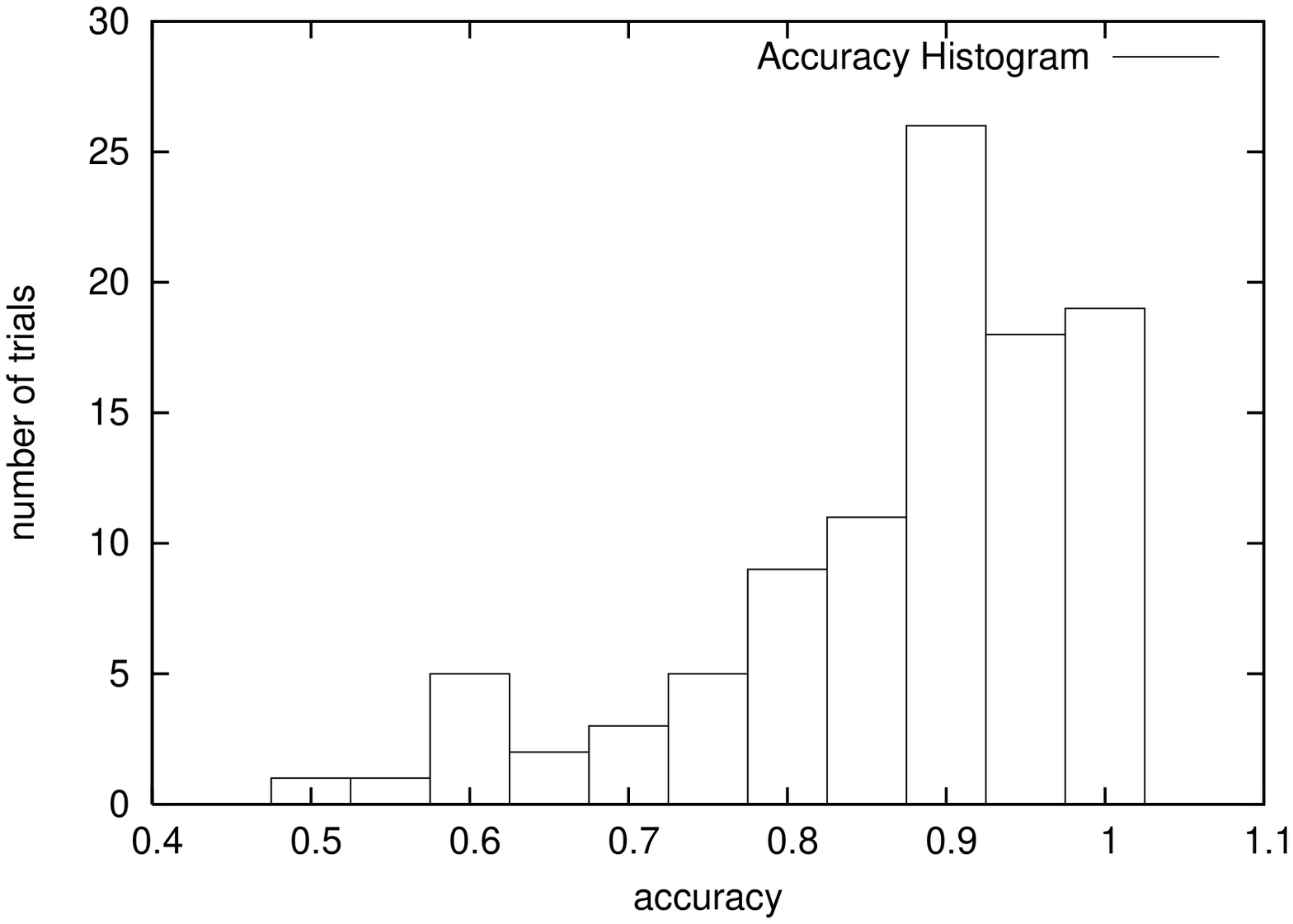}
}
\caption{Histogram of accuracies over 100 trials of WordNet experiment.}
\label{fig.wordnethisto}
\end{figure}
A histogram of agreement accuracies 
is shown in Figure~\ref{fig.wordnethisto}. 
On average, our method turns out to agree well with the WordNet semantic
concordance made by human experts.  The mean of the accuracies 
of agreements is 0.8725.
The variance is $\approx 0.01367$, which gives a standard deviation of
$\approx 0.1169$. Thus, it
is rare to find agreement less than 
75\%. The total number of Google searches involved in this
randomized automatic trial is upper bounded by $100 \times 70 \times 6 \times 3 =
126,000$. A considerable savings resulted from the fact
that we can re-use certain google counts.
For every new term, in computing its 6-dimensional vector, the \NGD computed
with respect to the six anchors requires the counts for the anchors which needs to be
computed only once for each experiment, the count of the new term which can be
computed once, and the count of the joint occurrence of the new term and 
each of the six anchors, which has to be computed in each case.
Altogether, this gives a total of $6+70+70 \times 6 = 496$ for every experiment,
so $49,600$ google searches for the entire trial.

It is conceivable that other scores instead of the \NGD used in the
construction of 6-dimensional vectors work competetively.
Yet, something simple like ``the number of words used in common in their
dictionary definition'' (Google indexes dictionaries too) is
begging the question and unlikely to be successful. In \cite{malivitch:simmet}
the \NCD abbroach, compression of the literal objects, was compared
with a number of alternative approaches like the Euclidean distance
between frequency vectors of blocks. The alternatives gave results
that were completely unacceptable. In the current setting, we can
conceive of Euclidean vectors of word frequencies in the set of pages
corresponding to the search term. Apart from the fact that Google
does not support automatical analysis of all pages reported for
a search term, it would be computationally infeasible to analyze
the millions of pages involved. Thus, a competetive nontrivial alternative
to compare the present technique against is an interesting open question.

\section{Conclusion}
A comparison can be made with the {\em Cyc} project
\cite{cyc:intro}.  Cyc, a project of the commercial venture Cycorp, tries to
create artificial common sense.  Cyc's knowledge base consists of hundreds of
microtheories and hundreds of thousands of terms, as well as over a million
hand-crafted assertions written in a formal language called CycL~\cite{cyc:onto}.  CycL is an enhanced variety of first-order predicate logic.
This knowledge base was created over the course of decades by paid human
experts.  It is therefore of extremely high quality.  Google, on the other
hand, is almost completely unstructured, and offers only a primitive query
capability that is not nearly flexible enough to represent formal deduction.
But what it lacks in expressiveness Google makes up for in size; Google has
already indexed more than eight billion pages and shows no signs of slowing down.

\section*{Acknowledgment}
We thank the referees and others for comments on presentation.

\section{Appendix: Relation to LSA}\label{app.LSA}
The basis assumption of Latent Semantic Analysis 
is that ``the cognitive similarity
between any two words is reflected in the way they co-occur in small subsamples
of the language.'' In particular, this is implemented by constructing a matrix
with rows labeled by the $d$ documents involved,
and the columns labeled by the $a$ attributes (words, phrases).
The entries are the number of times the column attribute occurs in the 
row document. The entries are then processed by taking the logarithm of the entry
and dividing it by the number of documents the attribute occurred in, or some
other normalizing function. This results in a sparse but high-dimensional
matrix $A$. A main feature of LSA is to reduce the dimensionality of the matrix by
projecting it into an adequate subspace of lower dimension using singular value
decomposition $A= U D V^T$ where $U,V$ are orthogonal matrices and $D$ is
a diagonal matrix. The diagonal elements $\lambda_1 , \ldots , \lambda_p$
($p = \min \{d,a\}$) satisfy $\lambda_1 \geq \cdots \geq \lambda_p$,
and the closest matrix $A_k$ of dimension $k < \mbox{\rm Rank}(A)$
in terms of the so-called Frobenius norm is obtained by setting $\lambda_i =0$
for $i>k$. Using $A_k$ corresponds to using the most important dimensions.
Each attribute is now taken to correspond to a column vector in $A_k$,
and the similarity between two attributes is usually 
taken to be the cosine between their
two vectors. To compare LSA to our proposed method, 
the documents could be the web pages, the entries in matrix $A$ are
the frequencies of a search terms in each web page. This is then converted
as above to obtain vectors for each search term. Subsequently, the cosine
between vectors gives the similarity between the terms. 
LSA has been used
in a plethora of applications ranging from data base query systems to
synonymy answering systems in TOEFL tests. Comparing its performance to
our method is problematic for several reasons. First, 
the numerical quantity measuring the semantic distance between
pairs of terms cannot directly be compared, since they have quite different
epistimologies. Indirect comparison could be given using the method
as basis for a particular application, and comparing accuracies. However,  
application of LSA in terms 
of the web using Google is computationally out of the question, because the
matrix $A$ would have $10^{10}$ rows, even if
Google would report frequencies of occurrences in web pages and
identify the web pages properly. One would need to retrieve
the entire Google data base, which is many terabytes.
Moreover, as noted in Section~\ref{sect.materials}, each Google search takes
a significant amount of time, and we cannot automatically make more than a 
certain number of them per day.
An alternative
interpretation by considering the web
as a single document makes the matrix $A$ above into a vector 
and appears to defeat the LSA process altogether. 
Summarizing, the basic idea of our method is similar
to that of LSA in spirit. What is novel is that we can do it with
selected terms over a very large document collection, whereas LSA
involves matrix operations over a closed collection of limited size,
and hence is not possible to apply in the web context.


\section{Biographies of the Authors}

{\sc Rudi Cilibrasi} received his B.S. with honors
 from the California Institute of
Technology in 1996.  He has programmed computers for over two decades,
both in academia,
 and industry with various companies in Silicon Valley, including Microsoft, in
diverse areas such as machine learning, data compression, process control,
VLSI design, computer
graphics, computer security,
 and networking
protocols.
He is now a PhD student at the Centre for 
Mathematics and Computer Science (CWI)
 in the Netherlands, and expects to receive his PhD soon on the circle
of ideas of which this paper is representative.
He helped create the first publicly downloadable Normalized Compression/Google
Distance software, and is maintaining http://www.complearn.org now.
Home page: http://www.cwi.nl/$\sim$cilibrar/

{\sc Paul M.B. Vit\'anyi} is a Fellow
of the Centre for Mathematics and Computer Science (CWI)
in Amsterdam and is Professor of Computer Science
at the University of Amsterdam.  He serves on the editorial boards
of Distributed Computing (until 2003), Information Processing Letters,
Theory of Computing Systems, Parallel Processing Letters,
International journal of Foundations of Computer Science,
Journal of Computer and Systems Sciences (guest editor),
and elsewhere. He has worked on cellular automata,
computational complexity, distributed and parallel computing,
machine learning and prediction, physics of computation,
Kolmogorov complexity, quantum computing. Together with Ming Li
they pioneered applications of Kolmogorov complexity
and co-authored ``An Introduction to Kolmogorov Complexity
and its Applications,'' Springer-Verlag, New York, 1993 (2nd Edition 1997),
parts of which have been translated into Chinese,  Russian and Japanese.
Home page: http://www.cwi.nl/$\sim$paulv/

\end{document}